%% file: acl2020.tex
\documentclass[11pt,a4paper]{article}
\usepackage[hyperref]{acl2020}
\usepackage{times}
\usepackage{float}
\usepackage{amsmath}
\usepackage{graphicx}
\usepackage{booktabs}
\usepackage{latexsym}
\usepackage{url}
\usepackage{multirow}
\usepackage{caption}
\usepackage{subcaption}
\usepackage{microtype}
\usepackage{url}
\aclfinalcopy % Uncomment this line for the final submission
\def\aclpaperid{27} %  Enter the acl Paper ID here
\newcommand{\ignore}[1]{}

\newcommand{\smallpar}[1]{\vspace{-3pt}\paragraph{#1}}

\title{On-The-Fly Information Retrieval Augmentation for Language Models}

\author{
 Hai Wang ~~ David McAllester \\
 Toyota Technological Institute at Chicago, Chicago, IL, USA \\
  \{haiwang,mcallester\}@ttic.edu \\
}

\date{}

\begin{document}
\maketitle

% first check the story 
\input{abstract} % abstract 
\input{intro} % introduction
\input{related} % related work
\input{model} % model
\input{experiment} % experiment
\input{conclusion} % conclusion

\bibliography{anthology,acl2020}
\bibliographystyle{acl_natbib}

\end{document}

%% file: abstract.tex
\begin{abstract}

Here we experiment with the use of information retrieval as an augmentation for pre-trained language models. The text corpus used in information retrieval can be viewed as form of episodic memory which grows over time. By augmenting GPT 2.0 with information retrieval we achieve a zero shot 15\% relative reduction in perplexity on Gigaword corpus without any re-training. We also validate our IR augmentation on an event co-reference task.
\end{abstract}

%% file: intro.tex
\section{Introduction}

% then newest now -- 04/18/2020

We are interested in exploring the value of long term
episodic memory in language modeling. For example, a language model can be used in January to assign a probability distribution over the statements that will appear in the newspaper in March.  But one month later, in February, the distribution over the predictions for March should be updated to take into account factual developments since the previous prediction. Long term episodic memory should be taken into account when assigning a probability to a statement.

Here we take a simple approach in which a pre-trained GPT language model~\cite{radford2018improving,radford2019language} is zero-shot augmented with an episodic memory consisting simply of a corpus of past news articles. Conceptually the past news articles are viewed as additional training data
which can be legitimately accessed when evaluating on future text. In our most basic experiment we calculate the probability of a future article by first calculating the probability of its first $k$ sentences using the pre-trained GPT model. We then use the first $k$ sentences as a query in an information retrieval system to extract a relevant past article. We then insert the past article following the first $k$ sentences when calculating the probability of the remainder of the future article using the same pre-trained GPT model. This is a zero-shot augmentation in the sense that there is no additional training or fine tuning of the pre-trained model. Our results show that this augmentation significantly reduces perplexity. We also present various other experiments including results on fine-tuning the model in the presence of the memory and the effect of this memory on event co-reference.

%% file: related.tex
\section{Related Work}

Various language models have utilized external knowledge or long contexts ~\cite{paperno2016lambada,yang2017leveraging,PengNiRo19,khandelwal2018sharp,ghosh2016contextual,lau2017topically,grave2016improving,parthasarathi2018extending}. But these papers do not address the question of whether additional context or external knowledge is useful as a zero-shot augmentation of large scale pre-trained NLP models.

The value of external knowledge has previously been demonstrated for NLP tasks such as natural language inference~\cite{chen2018neural,yang2019enhancing}, language generation~\cite{parthasarathi2018extending}, knowledge base completion~\cite{toutanova2015representing,das2017go} and question answering~\cite{sun2019pullnet,sun2018open,dhingra2017linguistic}. However, all those prior works assume the model is small and trained from scratch.

As large scale pre-trained models have become more powerful it is not immediately clear whether external resources can still add value. The only work we know of on using external resources in modern large scale models is~\citet{yang2019enhancing} where a human curated external lexical resource is used to improve BERT.

 Our approach bears some resemblance to
 neural cache models~\cite{grave2016improving}. However, neural cache models store past hidden states as memory and accesses them through a dot product with the current hidden states. This is different
 from retrieving knowledge from a corpus-sized
 memory.

\ignore{
Great success has been achieved by first pre-training deep neural networks and then fine-tuning them on downstream tasks~\cite{radford2018improving,devlin2018bert,radford2019language}. 
%It significantly outperforms methods that use pre-trained language models as auxiliary features~\cite{peters2018deep}. 
Instead of using LSTM~\cite{peters2018deep}, transformer~\cite{vaswani2017attention} has been a popular choice since they can capture long range structures. However, it's time-consuming and resource-intensive to pre-train a deep model such as GPT 2.0~\cite{radford2019language} and BERT~\cite{devlin2018bert}. It's also unclear whether the pre-trained models already learnt sufficient knowledge. 
}

Our approach is also somewhat related to memory networks~\cite{weston2014memory}. Memory networks have a memory module which can be learnt jointly with other components. It has shown success in applications such as machine reading comprehension~\cite{kumar2016ask,xiong2016dynamic,shi2016hierarchical} and visual question answering~\cite{na2017read,ma2018visual,su2018learning}. Significant progress in memory networks has been achieved in both architecture ~\cite{chandar2016hierarchical,miller2016key,gulcehre2017memory} and model scale~\cite{rae2016scaling, guillaume2019large}.

Several papers have formulated, and experimented with, scalable memory networks --- memory networks that employ some method of efficiently reading and writing to very large neural memories.  This is done with approximate nearest neighbor methods in~\citet{rae2016scaling} and with product keys in~\citet{guillaume2019large}. These large memories are used to provide additional model capacity where the memory contents are trained over a large data set
using gradient descent training, just as one would train the parameters of a very large network.  It is shown in~\citet{guillaume2019large} that it is possible to insert a large memory as a layer in a transformer architecture resulting a model where the same number of parameters and the same performance can be achieved with half the layers and with much faster training time than a standard transformer architecture.  Here, however, we are proposing zero-shot augmentation with an external data source used as an episodic memory.

The use of key-value memories in~\citet{miller2016key} is particularly similar to our model.  Key-value memories were used there in treating a corpus of Wikipedia movie pages as a memory for answering questions about movies. As in our system, articles were extracted using word based information retrieval. Each article was encoded as a vector which was then given to a question answering architecture. This was shown to improve on automated knowledge base extraction from the same corpus but was still not competitive with human curated knowledge graphs for movies. Here we give the text of the retrieved article directly to the language model architecture and focus on augmenting large scale language models.

%% file: model.tex
\section{Model}
 
\begin{figure}[!htbp]
\centering
\includegraphics[scale=0.4]{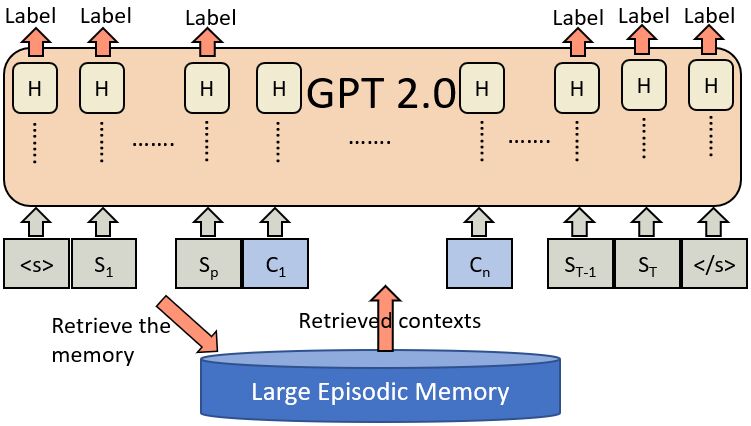}
\caption{GPT with large episodic memory component}
\label{fig:overview_model}
\end{figure}

We use the pre-trained transformer GPT 2.0~\cite{radford2019language}. Let $W_{w}$ and $W_{p}$ be the subword and position embeddings respectively. Let $M$ denote the total number of layers, for a token at time step $t$, the $m$-th layer's hidden state $h_{t}^{m}$ is given by:
\begin{equation}
\nonumber
%\small
h_{t}^{m} = \begin{cases}
%h^m = \begin{cases}
W_{w} + W_{p} &\text{if $m=0$}\\
\text{TB}(h_{t}^{m-1}) & \text{if $1\leq m \leq M$}
%\text{TB}(h^{m-1})  &\text{if $1\leq m \leq M$}
\end{cases}
\end{equation}

where TB stands for Transformer Block.
%, which is a module containing a MLP residual connection~\cite{he2016deep} and LayerNorm~\cite{ba2016layer}. 
We use last layer's hidden state $h_{t}^{M}$ as the presentation $H_{t}$ for the token at time step $t$. We augment GPT 2.0 with a large episodic memory component, and the overall architecture is shown in Figure \ref{fig:overview_model}. 

For a sequence $S$ with $T$ tokens, let $S_1$, $\ldots$, $S_p$ be the tokens of the first $k$ sentences. Let $C$ be a sequence (article) retrieved from memory using the first $k$ sentences as the query, the vector $H_{t}$ is:
\begin{equation}
\nonumber
%\small
H_{t} = \begin{cases}
\text{GPT}(S_{1},\dots, S_{t}),\text{if $t \leq p$}& \\
\text{GPT}(S_{1},\dots,S_p,C,\ldots, S_{t}),\text{otherwise}&
\end{cases}
\end{equation}

That's to say, for the first $k$ sentences, we directly feed them to GPT to obtain their representations. For remaining sentences, their representations are conditioned on both the first $k$ sentences and the retrieved context $C$. Table \ref{tab:comparision} compares features of our simple memory augmentation with those of other memory models.

\begin{table}[!htbp]
\begin{center}
\begin{tabular}{lccc}
\bf Model & \bf episodic & \bf search & \bf memory size \\
\toprule
 DMN & yes  & exact & $\sim$1K words \\
 SAM: & no & approx  & $\sim$100K slots \\
 KVM: & yes  & exact &  $\leq$ 1M slots \\
 LMN: & no  & exact & $\sim$1M slots \\
 Ours: & yes  & approx & $\sim$10M documents \\
\end{tabular}
\caption{Comparison between different models. DMN: Dynamic Memory Network~\cite{xiong2016dynamic}; SAM: Sparse Access Memory~\cite{rae2016scaling}; KVM: Key Value Memory~\cite{miller2016key}; LMN: Large Memory Network~\cite{guillaume2019large}. Memory size is measured in their own words.}
\label{tab:comparision}
\end{center}
\end{table}

%% file: experiment.tex
\section{Experiments}

 We focus on two tasks: document level language modelling and event co-retrieved . In both tasks we take a document as input and use first $k$ sentences to query the memory. To calculate the perplexity of a document, we compute the log-probability of a document by multiplying byte level probability, then divide the log-probability by the actual word count in the \textit{query} document. 
 
We use Gigaword~\cite{parker2011english} as both our language modeling
test set and as our external memory. Gigaword contains news from different sources such as NY Times and XinHua News etc. For language modelling we use the NY Times portion because it is written by native English speakers. Since GPT 2.0 is trained on Common Crawl
%\footnote{https://commoncrawl.org/} 
which contains news collections started from 2008. To avoid testing on GPT-2 training data, we use Gigaword articles collected prior to 2008. For the pre-trained language model we use GPT 2.0~\cite{radford2019language} \footnote{https://github.com/huggingface/pytorch-transformers}. It contains three pre-trained models: GPT Small, Medium and Large.

For information retrieval we use Lucene due to its simplicity. %\footnote{https://lucene.apache.org}. 
Given a query document we first do sentence and word tokenization and then use the first $k$ sentences to retrieve top 20 retrieved documents with the default TF-IDF distance metric provided by Lucene. Since too distant document pairs are uninformative and too related document pairs tends to be duplicates of the test article, we further filter those top ranked documents by time stamp, news source and cosine similarity. More specifically, we choose the highest ranked retrieved  document that simultaneously satisfies the following three conditions: it comes from a different news source; it appears earlier but within two weeks time window of the test document, and the bag of word cosine similarity between the test and the retrieved  cannot be larger than $0.6\alpha$ where $\alpha$ is the largest bag of word cosine similarity between the test article and any retrieved articles. To support fine-tuning experiments we constructed
a corpus of pairs of a \textit{query} article and a cached \textit{retrieved} document. We split the dataset into train/dev/test by query document's time stamp. The train/dev/test size is: 79622,16927,8045. For zero-shot experiments we use the test set of 8045 articles.  We do experiments with $k \in \{1,2,5\}$.

To check the quality of \textit{query-retrieved } pairs, we randomly sample 100 pairs from dev set and compute the bag of word cosine similarity between the two documents. The mean cosine similarity is 0.15. We also manually inspect them: we ask two NLP researchers to annotate the \textit{query-retrieved } pair as ``BAD" or ``OK" independently, i.e., if two documents are almost duplicates or totally unrelated, then it's ``BAD", otherwise, it's ``OK". Among 100 pairs, 83 pairs are ``OK", 17 pairs are ``BAD" due to irrelevance. The Cohen's kappa coefficient between two annotations is 0.94. 

%\vspace{-1.1em}
\subsection{Language modelling}
\label{exp:result:lm}
%\vspace{-0.1em}

For language modeling we try zero-shot memory augmentation, fine-tuned memory augmentation, and training a small memory-augmented network from scratch. When training, we use the Adam optimizer from GPT 1.0~\cite{radfordimproving}. The learning rate is 0.001, weight decay parameter is 0.01, the warm up proportion is 0.1. For other parameters, we use the default values from GPT 2.0. The fine-tuning on Gigaword takes less than one day with a single GPU.

\smallpar{Zero-shot and fine-tuning results} Following ~\citet{radford2019language}, we first evaluate our model on Gigaword with zero-shot setting and then fine-tune the model. The results are given in Table \ref{result:zero-shot-giga}.  

\begin{table}[H]
\begin{center}
\begin{tabular}{lllll}
\multicolumn{1}{l}{\bf Model Size}  & \multicolumn{1}{c}{\bf woc} & \multicolumn{1}{c}{\bf k=1} & \multicolumn{1}{c}{\bf k=2} & \multicolumn{1}{c}{\bf k=5} \\
\midrule
%\multicolumn{5}{l}{Zero shot} \\
%\midrule
GPT-Small         & 35.15 & 29.29 & 30.54 & 32.38 \\
GPT-Medium       & 22.78 & 19.84 & 20.54 & 21.48  \\
GPT-Large            & 19.90 & 17.41 & 18.00 & 18.80 \\
%\midrule
%\multicolumn{5}{l}{Fine tune} \\
\midrule
GPT-Small         & 23.03 & 21.01 & 21.89 & 22.66 \\
%GPT-Medium             & & & & \\
%GPT-Large            & & & &  \\
\end{tabular}
\end{center}
\caption{Perplexity for zero-shot (top 3 rows) and fine-tuning (last row) settings when use different $k$ to retrieve the context. \textbf{woc}: without retrieved context.}
\label{result:zero-shot-giga}
\end{table}
%\vspace{-0.5pt}

From Table \ref{result:zero-shot-giga}, we see that with additional context retrieved from episodic memory, for all different GPT models, we obtain significantly lower perplexity than using original GPT 2.0. When fine tuning the model with context, we can further reduce the overall perplexity. We only fine tune GPT small due to our GPU memory constraints. Preliminary analysis indicates that most of the perplexity reduction comes at content words and semantically rich words where predictions require broader context. This is consistent with the phenomena found in~\citet{khandelwal2018sharp}. We further find that smaller $k$ leads to slightly worse retrieval quality, however, more continued sentences will benefit from the retrieved context. Since Gigaword contains newswire, the first several sentences usually are importation summarizations, thus overall, smaller $k$ will result in lower perplexity.     

\smallpar{Train from scratch} We also investigate training this form of memory-augmented model from scratch on our query-retrieved pairs. For these experiments we train smaller transformers and the results are given in Table \ref{result:train-giga}. From Table \ref{result:train-giga}, we see that additional context still helps and we can get decent perplexity even with quite small models.

\begin{table}[H]
\begin{center}
\begin{tabular}{lllll}
\multicolumn{1}{l}{\bf Model Config}  & \multicolumn{1}{c}{\bf woc} & \multicolumn{1}{c}{\bf k=1} & \multicolumn{1}{c}{\bf k=2} & \multicolumn{1}{c}{\bf k=5} \\
\midrule
%train, test          & & & \\
%train, test              & & & \\
E=384,H=6,L=6             & 35.62 & 31.94 & 33.18 & 35.26 \\
%train, test             & & & \\
%\midrule
%train, test          & & & \\
%train, test              & & & \\
E=384,H=8,L=8             & 33.67 & 29.62 & 30.76 & 32.73 \\
%train, test             & & & \\
%\midrule
%train, test          & & & \\
%train, test              & & & \\
E=576,H=8,L=8             & 31.32 & 27.38 & 28.54 & 30.63 \\
\end{tabular}
\end{center}
\caption{Perplexity when train from scratch. E: hidden states dimensionality; H: \# of head; L: \# of layer. GPT-Small has the configuration: E=764, H=12, L=12.}
\label{result:train-giga}
\end{table}

\smallpar{When context is irrelevant} We also evaluate our method on Wikitext-2/103, in which the retrieved context is irrelevant due to domain difference between Wikipedia and Gigaword. In this case, we use the most top ranked document from Gigaword as reference. Table \ref{result:zero-shot-wikitext} shows that irrelevant contexts
have very little impact on perplexity.

\begin{table}[H]
\begin{center}
\begin{tabular}{lllll}
\multicolumn{1}{l}{\bf Dataset}  & \multicolumn{1}{c}{\bf woc} & \multicolumn{1}{c}{\bf k=1} & \multicolumn{1}{c}{\bf k=2} & \multicolumn{1}{c}{\bf k=5} \\
\midrule
Wikitext-2         & 28.67 & 28.96 & 28.95 & 28.70 \\
Wikitext-103         &25.38  & 25.68 & 25.56 & 25.39 \\
\end{tabular}
\end{center}
%\vspace{-2pt}
\caption{Zero-shot perplexity using GPT-Small}
\label{result:zero-shot-wikitext}
\end{table}

\subsection{Event Co-reference}

Intuitively episodic memory is useful because it contains information about the particular events mentioned in the test document. 
%Therefore we might expect the attention over the retrieved  document to be informative for event co-reference.
With this in mind we evaluate our approach on the event co-reference dataset ECB+~\cite{cybulska2014using}. ECB+ contains 982 documents clustered into 43 topics, and has two evaluation settings: coreferring mentions occurring within a single document (within document) or across a document collection (cross document). For the event co-reference pipeline, we follow the joint modeling method of~\citet{barhom2019revisiting} where they jointly represented entity and event mentions with various features and learned a pairwise mention/entity scorer for coreference classification. We augment their mention features with the mention's vector representations extracted from either GPT 2.0 or our zero-shot augmented GPT 2.0. For event co-reference, we use the whole test document to retrieve the context from Gigaword. From Table \ref{result:event-coref}, we see that the context can help boost the CONLL F1 score. 

%\cite{minard2016meantime}, \cite{o2016richer}, \cite{clark2019does} \cite{jawahar2019does}, survey paper: \cite{lu2018event}
%baselines: \cite{yang2015hierarchical}, \cite{lee2012joint}, \cite{barhom2019revisiting}
\begin{table}[!htbp]
\begin{center}
\begin{tabular}{lccc}
\multicolumn{1}{l}{System} & \multicolumn{1}{c}{MUC} & \multicolumn{1}{c}{ $\text{B}^{3}$ } & \multicolumn{1}{c}{CONLL} \\
\toprule
\multicolumn{4}{l}{Within Document} \\
\midrule
% \cite{yang2015hierarchical}     & 53.4 & 75.4  & 66.8 \\
% CV   &  &  &  \\
KCP   & 63.0 & 92.0 & 81.0 \\
JM & 70.9 & 93.5 & 85.1 \\
JM+GPT & 80.1 & 93.5 & 85.2 \\
$\text{JM+GPT+CTX}^{\clubsuit}$ & 80.2 & 93.9 & 85.4 \\
%GPT-2.0       & 38.2 &  &  \\
%GPT-2.0+context       & 38.5 & & \\
%small     & 34.3 & &  \\
%small+context      & 34.8 & &  \\
%cos-GPT-2.0       & 38.6 &  &   \\
%cos-GPT-2.0+context       & 38.9 & &  \\
%cos-small     & 34.7 & &   \\
%cos-small+context      & 35.3 & & \\
\midrule
\multicolumn{4}{l}{Combined Within and Cross Document} \\
\midrule
%\cite{lee2012joint}  &  62.7 & 67.7 &  54.8 \\
%\cite{yang2015hierarchical}     & 73.1 & 53.5 &  58.7 \\
CV   & 73.0 & 74.0 & 73.0 \\
KCP     & 69.0 & 69.0 &  69.0 \\
JM & 80.9 & 80.3 & 79.5 \\
JM+GPT & 81.2 & 80.2 & 79.6 \\
$\text{JM+GPT+CTX}^{\clubsuit}$ & 81.3 & 80.5 & 79.8 \\
%GPT-2.0        & &   & \\
%GPT-2.0+context      &  & & \\
%small     & & &  \\
%small+context     & & & \\
\end{tabular}
\end{center}

\caption{F1 score on ECB+ dataset. KCP:~\citet{kenyon2018resolving} where they add a clustering-oriented regularization term;
CV:~\citet{cybulska2015translating} where they add the feature calculated from ``event template"; JM:~\citet{barhom2019revisiting}.
$\clubsuit$: we also feed the retrieved context to GPT to get the representation.}
\label{result:event-coref}
\end{table}
% CONLL F1 is not statistical significant

%% file: conclusion.tex
\section{Conclusion}

In this paper we propose a method to augment a pre-trained NLP model with a large episodic memory. Unlike previous work, we use information retrieval to handle a large external
corpus of text and feed retrieved documents directly to language models.
Evaluation results on language modelling and event co-reference show the promise of our method. To the best of our knowledge, this is the first work that augments pre-trained NLP models with large episodic memory. In principle, the memory-augmented GPT-2 can be used as a variant of GPT-2 for any downstream tasks, such as GLUE tasks~\cite{wang2018glue}, although we have not experimented with that here.

%% file: acl2020.bbl
\begin{thebibliography}{36}
\expandafter\ifx\csname natexlab\endcsname\relax\def\natexlab#1{#1}\fi

\bibitem[{Barhom et~al.(2019)Barhom, Shwartz, Eirew, Bugert, Reimers, and
  Dagan}]{barhom2019revisiting}
Shany Barhom, Vered Shwartz, Alon Eirew, Michael Bugert, Nils Reimers, and Ido
  Dagan. 2019.
\newblock \href {https://www.aclweb.org/anthology/P19-1409.pdf} {Revisiting
  joint modeling of cross-document entity and event coreference resolution}.
\newblock pages 4179--4189.

\bibitem[{Chandar et~al.(2016)Chandar, Ahn, Larochelle, Vincent, Tesauro, and
  Bengio}]{chandar2016hierarchical}
Sarath Chandar, Sungjin Ahn, Hugo Larochelle, Pascal Vincent, Gerald Tesauro,
  and Yoshua Bengio. 2016.
\newblock \href {https://www.aclweb.org/anthology/C16-1216.pdf} {Hierarchical
  memory networks}.
\newblock \emph{arXiv preprint arXiv:1605.07427}.

\bibitem[{Chen et~al.(2018)Chen, Zhu, Ling, Inkpen, and Wei}]{chen2018neural}
Qian Chen, Xiaodan Zhu, Zhen-Hua Ling, Diana Inkpen, and Si~Wei. 2018.
\newblock \href {https://www.aclweb.org/anthology/P18-1224.pdf} {Neural natural
  language inference models enhanced with external knowledge}.
\newblock In \emph{Proceedings of the 56th Annual Meeting of the Association
  for Computational Linguistics (Volume 1: Long Papers)}, pages 2406--2417.

\bibitem[{Cybulska and Vossen(2014)}]{cybulska2014using}
Agata Cybulska and Piek Vossen. 2014.
\newblock \href
  {https://pdfs.semanticscholar.org/0fab/eb29eee19ca80b6f424d8cd86ac52ac96eb0.pdf}
  {Using a sledgehammer to crack a nut? lexical diversity and event coreference
  resolution}.
\newblock In \emph{Proceedings of the Ninth International Conference on
  Language Resources and Evaluation (LREC-2014)}, pages 4545--4552.

\bibitem[{Cybulska and Vossen(2015)}]{cybulska2015translating}
Agata Cybulska and Piek Vossen. 2015.
\newblock \href {https://www.aclweb.org/anthology/W15-0801.pdf} {Translating
  granularity of event slots into features for event coreference resolution}.
\newblock In \emph{Proceedings of the the 3rd Workshop on EVENTS: Definition,
  Detection, Coreference, and Representation}, pages 1--10.

\bibitem[{Das et~al.(2017)Das, Dhuliawala, Zaheer, Vilnis, Durugkar,
  Krishnamurthy, Smola, and McCallum}]{das2017go}
Rajarshi Das, Shehzaad Dhuliawala, Manzil Zaheer, Luke Vilnis, Ishan Durugkar,
  Akshay Krishnamurthy, Alex Smola, and Andrew McCallum. 2017.
\newblock \href {https://arxiv.org/pdf/1711.05851.pdf} {Go for a walk and
  arrive at the answer: Reasoning over paths in knowledge bases using
  reinforcement learning}.
\newblock \emph{arXiv preprint arXiv:1711.05851}.

\bibitem[{Dhingra et~al.(2017)Dhingra, Yang, Cohen, and
  Salakhutdinov}]{dhingra2017linguistic}
Bhuwan Dhingra, Zhilin Yang, William~W Cohen, and Ruslan Salakhutdinov. 2017.
\newblock \href {https://arxiv.org/pdf/1703.02620.pdf} {Linguistic knowledge as
  memory for recurrent neural networks}.
\newblock \emph{arXiv preprint arXiv:1703.02620}.

\bibitem[{Ghosh et~al.(2016)Ghosh, Vinyals, Strope, Roy, Dean, and
  Heck}]{ghosh2016contextual}
Shalini Ghosh, Oriol Vinyals, Brian Strope, Scott Roy, Tom Dean, and Larry
  Heck. 2016.
\newblock \href {https://arxiv.org/pdf/1602.06291.pdf} {Contextual lstm (clstm)
  models for large scale nlp tasks}.
\newblock \emph{arXiv preprint arXiv:1602.06291}.

\bibitem[{Grave et~al.(2016)Grave, Joulin, and Usunier}]{grave2016improving}
Edouard Grave, Armand Joulin, and Nicolas Usunier. 2016.
\newblock \href {https://arxiv.org/pdf/1612.04426.pdf} {Improving neural
  language models with a continuous cache}.
\newblock \emph{arXiv preprint arXiv:1612.04426}.

\bibitem[{Gulcehre et~al.(2017)Gulcehre, Chandar, and
  Bengio}]{gulcehre2017memory}
Caglar Gulcehre, Sarath Chandar, and Yoshua Bengio. 2017.
\newblock \href {https://arxiv.org/pdf/1701.08718.pdf} {Memory augmented neural
  networks with wormhole connections}.
\newblock \emph{arXiv preprint arXiv:1701.08718}.

\bibitem[{Kenyon-Dean et~al.(2018)Kenyon-Dean, Cheung, and
  Precup}]{kenyon2018resolving}
Kian Kenyon-Dean, Jackie Chi~Kit Cheung, and Doina Precup. 2018.
\newblock \href {https://www.aclweb.org/anthology/S18-2001.pdf} {Resolving
  event coreference with supervised representation learning and
  clustering-oriented regularization}.
\newblock In \emph{Proceedings of the Seventh Joint Conference on Lexical and
  Computational Semantics}, pages 1--10.

\bibitem[{Khandelwal et~al.(2018)Khandelwal, He, Qi, and
  Jurafsky}]{khandelwal2018sharp}
Urvashi Khandelwal, He~He, Peng Qi, and Dan Jurafsky. 2018.
\newblock \href {https://www.aclweb.org/anthology/P18-1027.pdf} {Sharp nearby,
  fuzzy far away: How neural language models use context}.
\newblock In \emph{Proceedings of the 56th Annual Meeting of the Association
  for Computational Linguistics (Volume 1: Long Papers)}, pages 284--294.

\bibitem[{Kumar et~al.(2016{\natexlab{a}})Kumar, Irsoy, Ondruska, Iyyer,
  Bradbury, Gulrajani, Zhong, Paulus, and Socher}]{kumar2016ask}
Ankit Kumar, Ozan Irsoy, Peter Ondruska, Mohit Iyyer, James Bradbury, Ishaan
  Gulrajani, Victor Zhong, Romain Paulus, and Richard Socher.
  2016{\natexlab{a}}.
\newblock \href {https://arxiv.org/pdf/1506.07285.pdf} {Ask me anything:
  Dynamic memory networks for natural language processing}.
\newblock In \emph{International conference on machine learning}, pages
  1378--1387.

\bibitem[{Kumar et~al.(2016{\natexlab{b}})Kumar, Irsoy, Ondruska, Iyyer,
  Bradbury, Gulrajani, Zhong, Paulus, and Socher}]{xiong2016dynamic}
Ankit Kumar, Ozan Irsoy, Peter Ondruska, Mohit Iyyer, James Bradbury, Ishaan
  Gulrajani, Victor Zhong, Romain Paulus, and Richard Socher.
  2016{\natexlab{b}}.
\newblock \href {https://arxiv.org/pdf/1506.07285.pdf} {Ask me anything:
  Dynamic memory networks for natural language processing}.
\newblock In \emph{International conference on machine learning}, pages
  1378--1387.

\bibitem[{Lample et~al.(2019)Lample, Sablayrolles, Ranzato, Denoyer, and
  J{\'e}gou}]{guillaume2019large}
Guillaume Lample, Alexandre Sablayrolles, Marc'Aurelio Ranzato, Ludovic
  Denoyer, and Herv{\'e} J{\'e}gou. 2019.
\newblock \href {https://arxiv.org/pdf/1907.05242.pdf} {Large memory layers
  with product keys}.
\newblock In \emph{Advances in Neural Information Processing Systems}, pages
  8546--8557.

\bibitem[{Lau et~al.(2017)Lau, Baldwin, and Cohn}]{lau2017topically}
Jey~Han Lau, Timothy Baldwin, and Trevor Cohn. 2017.
\newblock \href {https://arxiv.org/pdf/1704.08012.pdf} {Topically driven neural
  language model}.
\newblock In \emph{Proceedings of the 55th Annual Meeting of the Association
  for Computational Linguistics (Volume 1: Long Papers)}, pages 355--365.

\bibitem[{Ma et~al.(2018)Ma, Shen, Dick, Wu, Wang, van~den Hengel, and
  Reid}]{ma2018visual}
Chao Ma, Chunhua Shen, Anthony Dick, Qi~Wu, Peng Wang, Anton van~den Hengel,
  and Ian Reid. 2018.
\newblock \href {https://arxiv.org/pdf/1707.04968.pdf} {Visual question
  answering with memory-augmented networks}.
\newblock In \emph{Proceedings of the IEEE Conference on Computer Vision and
  Pattern Recognition}, pages 6975--6984.

\bibitem[{Miller et~al.(2016)Miller, Fisch, Dodge, Karimi, Bordes, and
  Weston}]{miller2016key}
Alexander Miller, Adam Fisch, Jesse Dodge, Amir-Hossein Karimi, Antoine Bordes,
  and Jason Weston. 2016.
\newblock \href {https://arxiv.org/pdf/1606.03126.pdf} {Key-value memory
  networks for directly reading documents}.
\newblock In \emph{Proceedings of the 2016 Conference on Empirical Methods in
  Natural Language Processing}, pages 1400--1409.

\bibitem[{Na et~al.(2017)Na, Lee, Kim, and Kim}]{na2017read}
Seil Na, Sangho Lee, Jisung Kim, and Gunhee Kim. 2017.
\newblock \href {https://arxiv.org/pdf/1709.09345.pdf} {A read-write memory
  network for movie story understanding}.
\newblock In \emph{Proceedings of the IEEE International Conference on Computer
  Vision}, pages 677--685.

\bibitem[{Paperno et~al.(2016)Paperno, Kruszewski, Lazaridou, Pham, Bernardi,
  Pezzelle, Baroni, Boleda, and Fernandez}]{paperno2016lambada}
Denis Paperno, Germ{\'a}n Kruszewski, Angeliki Lazaridou, Ngoc~Quan Pham,
  Raffaella Bernardi, Sandro Pezzelle, Marco Baroni, Gemma Boleda, and Raquel
  Fernandez. 2016.
\newblock \href {https://arxiv.org/pdf/1606.06031.pdf} {The lambada dataset:
  Word prediction requiring a broad discourse context}.
\newblock In \emph{Proceedings of the 54th Annual Meeting of the Association
  for Computational Linguistics (Volume 1: Long Papers)}, pages 1525--1534.

\bibitem[{Parker et~al.(2011)Parker, Graff, Kong, Chen, and
  Maeda}]{parker2011english}
Robert Parker, David Graff, Junbo Kong, Ke~Chen, and Kazuaki Maeda. 2011.
\newblock \href {https://catalog.ldc.upenn.edu/LDC2003T05} {English gigaword}.
\newblock \emph{Linguistic Data Consortium}.

\bibitem[{Parthasarathi and Pineau(2018)}]{parthasarathi2018extending}
Prasanna Parthasarathi and Joelle Pineau. 2018.
\newblock \href {https://arxiv.org/pdf/1809.05524.pdf} {Extending neural
  generative conversational model using external knowledge sources}.
\newblock In \emph{Proceedings of the 2018 Conference on Empirical Methods in
  Natural Language Processing}, pages 690--695.

\bibitem[{Peng et~al.(2019)Peng, Ning, and Roth}]{PengNiRo19}
Haoruo Peng, Qiang Ning, and Dan Roth. 2019.
\newblock \href {https://cogcomp.seas.upenn.edu/papers/PengNiRo19.pdf}
  {{KnowSemLM: A Knowledge Infused Semantic Language Model}}.
\newblock In \emph{Proc. of the Conference on Computational Natural Language
  Learning (CoNLL)}.

\bibitem[{Radford et~al.(2018{\natexlab{a}})Radford, Narasimhan, Salimans, and
  Sutskever}]{radford2018improving}
Alec Radford, Karthik Narasimhan, Tim Salimans, and Ilya Sutskever.
  2018{\natexlab{a}}.
\newblock \href {https://s3-us-west-2.amazonaws.
  com/openai-assets/researchcovers/languageunsupervised/language understanding
  paper.pdf} {Improving language understanding by generative pre-training}.
\newblock \emph{OpenAI Blog}.

\bibitem[{Radford et~al.(2018{\natexlab{b}})Radford, Narasimhan, Salimans, and
  Sutskever}]{radfordimproving}
Alec Radford, Karthik Narasimhan, Tim Salimans, and Ilya Sutskever.
  2018{\natexlab{b}}.
\newblock \href
  {https://s3-us-west-2.amazonaws.com/openai-assets/research-covers/language-unsupervised/language_understanding_paper.pdf}
  {Improving language understanding by generative pre-training}.
\newblock In \emph{Preprint}.

\bibitem[{Radford et~al.(2019)Radford, Wu, Child, Luan, Amodei, and
  Sutskever}]{radford2019language}
Alec Radford, Jeffrey Wu, Rewon Child, David Luan, Dario Amodei, and Ilya
  Sutskever. 2019.
\newblock \href
  {https://d4mucfpksywv.cloudfront.net/better-language-models/language_models_are_unsupervised_multitask_learners.pdf}
  {Language models are unsupervised multitask learners}.
\newblock \emph{OpenAI Blog}, 1(8).

\bibitem[{Rae et~al.(2016)Rae, Hunt, Danihelka, Harley, Senior, Wayne, Graves,
  and Lillicrap}]{rae2016scaling}
Jack Rae, Jonathan~J Hunt, Ivo Danihelka, Timothy Harley, Andrew~W Senior,
  Gregory Wayne, Alex Graves, and Timothy Lillicrap. 2016.
\newblock \href {https://arxiv.org/pdf/1610.09027.pdf} {Scaling
  memory-augmented neural networks with sparse reads and writes}.
\newblock In \emph{Advances in Neural Information Processing Systems}, pages
  3621--3629.

\bibitem[{Shi et~al.(2016)Shi, Yao, Zheng, Xu et~al.}]{shi2016hierarchical}
Jing Shi, Yiqun Yao, Suncong Zheng, Bo~Xu, et~al. 2016.
\newblock \href {https://www.aclweb.org/anthology/C16-1216.pdf} {Hierarchical
  memory networks for answer selection on unknown words}.
\newblock In \emph{Proceedings of COLING 2016, the 26th International
  Conference on Computational Linguistics: Technical Papers}, pages 2290--2299.

\bibitem[{Su et~al.(2018)Su, Zhu, Dong, Cai, Chen, and Li}]{su2018learning}
Zhou Su, Chen Zhu, Yinpeng Dong, Dongqi Cai, Yurong Chen, and Jianguo Li. 2018.
\newblock \href {https://arxiv.org/pdf/1806.04860.pdf} {Learning visual
  knowledge memory networks for visual question answering}.
\newblock In \emph{Proceedings of the IEEE Conference on Computer Vision and
  Pattern Recognition}, pages 7736--7745.

\bibitem[{Sun et~al.(2019)Sun, Bedrax-Weiss, and Cohen}]{sun2019pullnet}
Haitian Sun, Tania Bedrax-Weiss, and William~W Cohen. 2019.
\newblock \href {https://arxiv.org/pdf/1904.09537.pdf} {Pullnet: Open domain
  question answering with iterative retrieval on knowledge bases and text}.
\newblock \emph{arXiv preprint arXiv:1904.09537}.

\bibitem[{Sun et~al.(2018)Sun, Dhingra, Zaheer, Mazaitis, Salakhutdinov, and
  Cohen}]{sun2018open}
Haitian Sun, Bhuwan Dhingra, Manzil Zaheer, Kathryn Mazaitis, Ruslan
  Salakhutdinov, and William Cohen. 2018.
\newblock \href {https://www.aclweb.org/anthology/D18-1455.pdf} {Open domain
  question answering using early fusion of knowledge bases and text}.
\newblock In \emph{Proceedings of the 2018 Conference on Empirical Methods in
  Natural Language Processing}, pages 4231--4242.

\bibitem[{Toutanova et~al.(2015)Toutanova, Chen, Pantel, Poon, Choudhury, and
  Gamon}]{toutanova2015representing}
Kristina Toutanova, Danqi Chen, Patrick Pantel, Hoifung Poon, Pallavi
  Choudhury, and Michael Gamon. 2015.
\newblock \href {https://www.cs.princeton.edu/~danqic/papers/emnlp2015.pdf}
  {Representing text for joint embedding of text and knowledge bases}.
\newblock In \emph{Proceedings of the 2015 Conference on Empirical Methods in
  Natural Language Processing}, pages 1499--1509.

\bibitem[{Wang et~al.(2018)Wang, Singh, Michael, Hill, Levy, and
  Bowman}]{wang2018glue}
Alex Wang, Amanpreet Singh, Julian Michael, Felix Hill, Omer Levy, and Samuel
  Bowman. 2018.
\newblock \href {https://arxiv.org/abs/1804.07461} {Glue: A multi-task
  benchmark and analysis platform for natural language understanding}.
\newblock In \emph{Proceedings of the 2018 EMNLP Workshop BlackboxNLP:
  Analyzing and Interpreting Neural Networks for NLP}, pages 353--355.

\bibitem[{Weston et~al.(2014)Weston, Chopra, and Bordes}]{weston2014memory}
Jason Weston, Sumit Chopra, and Antoine Bordes. 2014.
\newblock \href {https://arxiv.org/pdf/1410.3916.pdf} {Memory networks}.
\newblock \emph{arXiv preprint arXiv:1410.3916}.

\bibitem[{Yang and Mitchell(2017)}]{yang2017leveraging}
Bishan Yang and Tom Mitchell. 2017.
\newblock \href {https://www.cs.cmu.edu/~bishan/papers/kblstm_acl2017.pdf}
  {Leveraging knowledge bases in lstms for improving machine reading}.
\newblock In \emph{Proceedings of the 55th Annual Meeting of the Association
  for Computational Linguistics (Volume 1: Long Papers)}, pages 1436--1446.

\bibitem[{Yang et~al.(2019)Yang, Zhu, Zhao, Zhang, and
  Feng}]{yang2019enhancing}
Xiaoyu Yang, Xiaodan Zhu, Huasha Zhao, Qiong Zhang, and Yufei Feng. 2019.
\newblock \href
  {https://link.springer.com/chapter/10.1007/978-3-030-18305-9_38} {Enhancing
  unsupervised pretraining with external knowledge for natural language
  inference}.
\newblock In \emph{Canadian Conference on Artificial Intelligence}, pages
  413--419. Springer.

\end{thebibliography}
